\documentclass{article} % For LaTeX2e
\usepackage{iclr2027_conference,times}

\usepackage{amsmath,amsfonts,bm}

\def\eqref#1{equation~\ref{#1}}
\def\1{\bm{1}}

\DeclareMathAlphabet{\mathsfit}{\encodingdefault}{\sfdefault}{m}{sl}
\SetMathAlphabet{\mathsfit}{bold}{\encodingdefault}{\sfdefault}{bx}{n}

\usepackage{hyperref}
\usepackage{url}
\usepackage{graphicx}
\usepackage{booktabs}
\usepackage{amsmath}
\usepackage{multirow}
\graphicspath{{figures/}}
\title{When local gains fail to transfer: Frozen Earth-observation embeddings across wildfires}

\author{
Philipp Stark\thanks{ORCID: \url{https://orcid.org/0000-0001-6529-1142}. Lund University research profile: \url{https://portal.research.lu.se/en/persons/philipp-stark/}.} \\
Department of Human Geography \\
Lund University \\
Sölvegatan 10, SE-223 62 Lund, Sweden \\
\texttt{philipp.stark@keg.lu.se}
\And
Alexandros Sopasakis\thanks{ORCID: \url{https://orcid.org/0000-0001-9167-3590}. Lund University research profile: \url{https://portal.research.lu.se/en/persons/alexandros-sopasakis/}.} \\
Centre for Mathematical Sciences \\
Lund University \\
Sölvegatan 18, SE-221 00 Lund, Sweden \\
\texttt{alexandros.sopasakis@math.lth.se}
\And
Ola Hall\thanks{ORCID: \url{https://orcid.org/0000-0002-9231-4028}. Lund University research profile: \url{https://portal.research.lu.se/en/persons/ola-hall/}.} \\
Department of Human Geography \\
Lund University \\
Sölvegatan 10, SE-223 62 Lund, Sweden \\
\texttt{ola.hall@keg.lu.se}
}
\makeatletter
\iclrfinaltrue
\makeatother

\begin{document}

\maketitle

\begin{abstract}
Frozen Earth-observation embeddings are judged almost entirely by spatially blocked cross-validation inside one study region. We show that this number does not predict accuracy in a new region, we show why, and we show the one setting in which such a model does keep working, using a protocol that needs only a linear probe and labels one already has. The testbed is wildfire, with Copernicus burned-area maps of six fires in Greece and Spain and descriptors from the year before each fire, comparing TESSERA and AlphaEarth with ESA WorldCover classes and annual Sentinel-2 index summaries. Inside a fire, the embeddings identify the burned land 0.05 to 0.13 ROC AUC better than the index summaries, and repeated fold allocations, spatial buffers, a block bootstrap, and gradient-boosted trees leave that margin unchanged. On a fire in another region, they lose 0.15 to 0.18 AUC, and the index summaries lose 0.06, so the three end within a few hundredths of each other. The representation is not the cause. Eight labelled blocks from the new region restore the embedding advantage and give a higher AUC than 59,000 labelled pixels from other regions, and the weight vector fitted in one region is nearly orthogonal to the vector fitted in the others, so the part that carries across regions is small and low-dimensional. Forecasting within a region is a different matter. Fitted on a fire that burned in 2023 and applied to a fire twelve kilometres away that burned in 2024, where nothing used postdates the target fire, TESSERA reaches 0.772 AUC and loses 0.04 against a classifier fitted inside the 2024 fire, while classifiers fitted in other regions lose 0.09 to 0.18. A region with one mapped fire can therefore forecast susceptibility for later fires there; a region without one cannot borrow a model from elsewhere, and every evaluation of a frozen embedding should report a held-out region.
\end{abstract}

\section{Introduction}
Earth-observation foundation models produce per-pixel embeddings that summarise multi-sensor satellite time series into continuous vectors \citep{lacoste_geobench_2023,jakubik_geospatial_foundation_models_2023}. A large body of work shows that such embeddings serve as reusable features for downstream geospatial tasks \citep{wang_ssl_remote_sensing_review_2022,rolf_mosaiks_2021,jean_tile2vec_2019,tseng_presto_2023, Astrom2025Predicting}. Such products as TESSERA \citep{feng_tessera_2026_cvpr,ucam_eo_tessera_repository_2026} and AlphaEarth \citep{brown_alphaearth_2025,google_satellite_embedding_v1_2025} release their embeddings globally at 10\,m as fixed annual descriptors. The intended use is to keep the embedding fixed and train a task-specific classifier on top of it. While this use is straightforward, how this mode of use should be evaluated is less settled, with recent work highlighting the lack of standardized evaluation practices for geospatial foundation models \citep{corleyNoOneKnows2026}. In this regard, we want to bring two important evaluation criteria into focus: cross-regional and temporal generalization. The first is whether a classifier trained in one region maintains accuracy in another, a step toward a general task-specific model. The second is whether a classifier trained in one region keeps its accuracy there in later years, which is particularly relevant for developing region-specific forecasting models.

In this study, we focus on the task of distinguishing burned from unburned areas following wildfire events by separating the pixels that burned in a wildfire from pixels of the surrounding landscape that did not. To evaluate the cross-regional and temporal dimensions, high-risk wildfire regions with confirmed fire events were pre-selected and cross-validated against each other. For the temporal dimension, we selected wildfire regions where additional fires occurred in consecutive years.
Each fire event is treated as one evaluation unit with both burned and unburned pixels. We took the clean labels from the Copernicus Emergency Management Service (CEMS) Rapid Mapping product \citep{ec_jrc_cems_rapid_mapping_catalogue}, a commonly used and publicly available source that supports reproducibility. We use annual embeddings from TESSERA and AlphaEarth as descriptors, selecting the embedding from the year before each fire. We also added two conventional baselines: the ESA WorldCover land-cover class \citep{zanaga_esa_worldcover_2021_v200_2022} and annual summaries of three Sentinel-2 vegetation and moisture indices. For the cross-regional evaluation, we study four wildfires from 2023 in three regions, covering two seasons: Rhodes and Evros in Greece, Cáceres in mainland Spain, and Tenerife, a volcanic Atlantic island in Spain. For the temporal evaluation, we also consider two fires in Attica, Greece: Parnitha (2023) and Varnavas (2024), twelve kilometres away. This pair allows us to evaluate transfer one year forward within the same region. We evaluate the pre-fire descriptors using both linear probes and tree-based models.

Our evaluation strategy follows a robust design aimed at assessing generalization across fires, regions, and time. Every descriptor comes from the year before the fire, so no information from the fire itself enters the inputs. Each fire is treated as one evaluation unit, and transfer is evaluated using classifiers trained on entirely different fires. We further examine how much performance can be recovered from a small number of labels from a new fire and compare the weight vectors learned across fires. In the temporal evaluation, both the training labels and descriptors used for the 2024 fire existed before it started, providing a forecasting setting. We find substantially stronger transfer across years within the same landscape than across regions.
%Every descriptor comes from the year before the fire, so nothing from the fire itself enters the inputs. We count each fire as one case, not each pixel. When a fire is tested, the classifier has been fitted on other fires only; it is scored on the spatial folds used for that fire's own within-fire result, and the two are compared fold by fold. We furthermore measure how much a small number of labels from the new fire recovers, and we compare the weight vectors learned in different fires directly. We then add a fifth and a sixth fire, twelve kilometres apart in Attica, one burned in August 2023 and the other in August 2024. Everything the probe uses for the 2024 fire- its training labels and its descriptors- existed before that fire started, so this is a forecast. The probe fitted on the 2023 fire loses 0.04 AUC on the 2024 fire, and probes fitted in the four other regions lose 0.09 to 0.18 on the same fire. The embedding advantage therefore holds across a year inside one landscape and does not hold across landscapes.

With this research, we are contributing in the following way:
\vspace{-.2cm}
\begin{itemize}
    \item \textbf{Cross-regional generalization.} Frozen embeddings consistently outperform conventional descriptors within fires, improving ROC AUC over the Sentinel-2 index baseline by 0.05–0.13, with their combination performing best across all four fires (ROC AUC: 0.832- 0.975). This advantage largely disappears across regions, where a small amount of local supervision outperforms substantially larger training sets from other regions, indicating that the embeddings contain useful signal but that their relationship to wildfire occurrence is strongly region-specific.
    \item \textbf{Temporal generalization.} Two fires in Attica, twelve kilometres apart and one year apart, provide a within-region forecasting test. TESSERA retains much of its within-fire performance when transferred from the earlier to the later fire, while AlphaEarth shows a weaker local advantage. Combining the nearby fire with pooled remote fires gives the best transfer for both embeddings, suggesting that within-region performance can persist across years but that spatial and temporal proximity alone is not sufficient.
    \item \textbf{Evaluation protocol.} We introduce a robust protocol for assessing cross-regional and temporal generalization using existing labels and lightweight classifiers. It combines held-out-fire transfer, adaptation with limited local supervision, and comparison of the fitted weight vectors across fires.
    \item \textbf{Data and reproducibility.} We assemble six fires and about 120,000 labelled pixels across four descriptor sets, with fixed spatial fold assignments to ensure reproducible evaluation.
    %\item \textbf{A benchmark and two reproducibility warnings.} Six fires and 120,000 pixels linking CEMS burned-area polygons, WorldCover labels, Sentinel-2 index summaries and the TESSERA and AlphaEarth embeddings, with block-to-fold assignments saved so that every comparison uses identical folds. Appendix~\ref{app:pitfalls} reports a widely used grouped cross-validator that returns nearly identical folds for every seed, and nodata pixels stored as all-zero vectors that act as a class label.
\end{itemize}
    %\item Two reproducibility warnings, detailed in Appendix~\ref{app:pitfalls}. A widely used grouped cross-validator returns nearly identical folds for every seed, and nodata pixels stored as all-zero embedding vectors can act as a class label.

\section{Related work}
Burned-area mapping from satellite data is a mature field \citep{chuvieco_burned_area_review_2019,szpakowski_jensen_rs_fire_ecology_2019,pettinari_chuvieco_global_fuel_2016}. 
Yet what burns within a wildfire is heterogeneous, depending on vegetation, fuel characteristics, and fire conditions \citep{Castilla2016WhatBurns,Ferster2016WhatBurnsWhy,Kolden2012WildfireConsumption,Crowley2023WholeSystem}, and the extent to which burned and unburned areas can be distinguished depends on how the pre-fire landscape is represented. Categorical land-cover products reduce heterogeneous landscapes to discrete classes and can therefore yield different descriptions of the same burned area \citep{fisher_pixel_snare_1997,foody_landcover_accuracy_2002,comber_what_is_land_cover_2005}. In contrast, representation learning for remote sensing has produced task-agnostic embeddings \citep{jean_tile2vec_2019,rolf_mosaiks_2021,tseng_presto_2023,wang_ssl_remote_sensing_review_2022} and benchmarks that evaluate them with frozen heads \citep{lacoste_geobench_2023}. TESSERA \citep{feng_tessera_2026_cvpr} provides global annual embeddings from Sentinel-1 and Sentinel-2 time series, while AlphaEarth Foundations \citep{brown_alphaearth_2025} provides global annual embedding fields from optical, radar, and other sources through Earth Engine \citep{google_satellite_embedding_v1_2025}. Both provide fixed, analysis-ready representations for label-efficient downstream tasks. Recent work has begun to examine how well these representations transfer. \citet{zhuang_alphaearth_wildfire_2026} use AlphaEarth embeddings for wildfire susceptibility mapping in Victoria and report substantially better transfer to two nearby Australian regions than models based on physical variables. \citet{koko_alphaearth_tessera_lcz_2026} compare AlphaEarth and TESSERA with Sentinel-1/2 composites for local climate zone mapping across five Swiss cities and examine both cross-city and temporal transfer. Our evaluation differs in its focus on event-level generalization. Every descriptor predates the corresponding fire, each fire is treated as one evaluation unit, and held-out fires are evaluated using probes trained on different fires. This allows us to separate within-fire performance from cross-regional and temporal generalization. Our sampling follows used--available designs in ecology \citep{johnson_usage_availability_1980,northrup_use_availability_design_2013,barbet_massin_pseudo_absences_2012}, while the spatial validation follows \citet{roberts_cross_validation_structure_2017} and \citet{ploton_spatial_validation_2020}.

\section{Method}
\subsection{Data}

\subsubsection{Wildfire events}

We use four CEMS Rapid Mapping wildfire activations from 2023 (Table~\ref{tab:events}, Appendix~\ref{app:data}). EMSR675 on Rhodes and EMSR686 in Evros are the two large Greek fires of that summer \citep{copernicus_emsr675_2023,copernicus_emsr686_2023}, EMSR667 in C\'aceres, Spain, is a spring fire on granite uplands under oak and pine \citep{copernicus_emsr667_2023}, and EMSR685 on Tenerife is an August fire in Canary pine forest on volcanic terrain \citep{copernicus_emsr685_2023}. For each activation, we take the final delineation product as the mapped burned extent. All 4 fires occurred in 2023, so 2022 is the pre-fire year for every descriptor.

Two further activations serve one experiment only, in Section~\ref{sec:attica}. EMSR690 burned the southern slopes of Mount Parnitha in North Attica, Greece, in August 2023 \citep{copernicus_emsr690_2023}, and EMSR746 burned the hills around Varnavas, in the same part of Attica, in August 2024 \citep{copernicus_emsr746_2024}. The two mapped areas lie twelve kilometres apart in the same vegetation, climate and terrain, and the fires are one year apart. The descriptors of the Parnitha fire come from 2022 and those of the Varnavas fire from 2023, so each fire is again described by the year before it burned.

\subsubsection{Burned and control pixel selection}

The unit of analysis is the 10\,m TESSERA pixel. Burned pixels are centres inside the burned polygons after a 30\,m interior buffer. Control pixels are drawn from the activation area outside the polygons and outside a 500\,m exclusion buffer, so they represent the locally available landscape rather than the fire margin. We sample 5,000 burned and 15,000 control pixels per fire with a fixed seed, which gives 80,000 pixels for the four 2023 fires and 40,000 more for the Attica pair. After the exclusions described below, 79,056 pixels enter every analysis of the four fires, with the same comparison mask for every representation. Controls describe the land that was available around the fire, not an estimate of what would have burned under other conditions. Figure~\ref{fig:data-overview} (Appendix~\ref{app:data}) shows the two Greek fires. The Spanish fires are sampled identically.

\subsubsection{Four descriptors of each pixel}

\textbf{WorldCover.} The ESA WorldCover 2021 v200 class at the pixel, sampled nearest-neighbour and one-hot encoded \citep{zanaga_esa_worldcover_2021_v200_2022}.

\textbf{Sentinel-2 index summaries.} From all cloud-masked 2022 Level-2A observations (Scene Classification Layer classes 4, 5 and 6 kept) we compute NDVI, NDMI and NBR per observation and summarise each index by its median, interquartile range, 10th and 90th percentiles and valid-observation count, giving 15 features, with the 20\,m bands bilinearly resampled to 10\,m first. The three count features encode scene coverage rather than the surface, so we report results with and without them.

\textbf{TESSERA embeddings.} The 128-dimensional TESSERA v1.0 annual embedding for 2022 at the pixel \citep{feng_tessera_2026_cvpr,ucam_eo_tessera_repository_2026}. TESSERA stores embeddings as signed 8-bit integers together with one float32 scale per pixel, and the official loader reconstructs the embedding as the product. A per-pixel scale is a per-row multiplier that column standardisation does not remove, so we obtained the scale files for all 58 tiles, use the dequantised embedding throughout, and report the raw integers as a sensitivity variant. 940 control pixels, 902 of them permanent water, have a zero scale and an all-zero integer vector. These all-zero vectors occur only among the controls, so a probe can learn that a missing embedding means a control pixel, which is a fact about the sampling design and not about the land surface. We exclude these pixels from every analysis, which changes any within-fire AUC by at most 0.004.

\textbf{AlphaEarth embeddings.} The 64-dimensional AlphaEarth Foundations annual embedding for 2022 \citep{brown_alphaearth_2025}, read from the Earth Engine Satellite Embedding dataset \citep{google_satellite_embedding_v1_2025} at the same pixel centres, nearest pixel at the native 10\,m. It comes from a different group, with different inputs and training; its vectors are unit-length floats, so no quantisation question arises, and every sampled pixel has a value. It is a second, independently built frozen embedding, not a controlled variation of the first.

\subsection{Experimental design}

\subsubsection{Probe and within-fire evaluation}

All comparisons use $L_2$-regularised logistic regression with balanced class weights and default regularisation, implemented in scikit-learn \citep{pedregosa_scikit_learn_2011}. Continuous features are standardised using training-fold statistics. TESSERA is used in its dequantised form, with the raw integer representation as a sensitivity variant. All descriptors are fixed inputs; only the probe is fitted to the labels. Performance is measured by area under the ROC curve (AUC).

Pixels are grouped into 5\,km blocks on the ETRS89-LAEA grid, and five-fold cross-validation holds out complete blocks \citep{roberts_cross_validation_structure_2017,ploton_spatial_validation_2020}. With only 39--90 blocks per fire, the assignment of blocks to folds noticeably affects the result, so we evaluate 21 spatial allocations. Allocation 0 is produced by scikit-learn's \texttt{StratifiedGroupKFold} with its default seed. For the remaining 20 allocations, blocks are visited in random order and assigned to the fold with the fewest burned pixels so far. We do not use \texttt{shuffle=True}, because it produces little variation in the resulting folds (Section~\ref{sec:discussion}). We report the mean and standard deviation of the fold-averaged AUC across allocations, and differences between descriptors are always computed within the same fold before averaging.

Adjacent blocks touch, and the closest test pixel can be only 22--45\,m from a training pixel. As a sensitivity analysis, we therefore remove training pixels within 500\,m and 1\,km of any test pixel. Uncertainty within a fire is estimated by resampling complete blocks with replacement within each test fold using 2,000 bootstrap replicates, without refitting.

\subsubsection{Robustness checks}

To test whether the results depend on the linear probe, we rerun every descriptor on the same persisted folds using tuned logistic regression and LightGBM gradient-boosted trees \citep{ke_lightgbm_2017}, using the first five of the 21 spatial fold allocations (allocations 0--4). Hyperparameters are selected separately within the training portion of each outer fold using three-fold grouped cross-validation, with identical search spaces for every descriptor set (Appendix~\ref{app:extra}); the outer test fold is never used for model selection. We additionally test the raw versus dequantised TESSERA representation, nodata handling, and the 500\,m and 1\,km spatial buffers described above.

\subsubsection{Generalization to held-out fires}

The central generalization test asks what happens when a probe is applied to a fire it has never seen. We evaluate one-to-one transfer by fitting on one fire and testing on each of the other three (12 directed comparisons), and leave-one-fire-out transfer by fitting jointly on three fires and testing on the fourth (four comparisons). Each target fire is scored on the same spatial folds and 21 allocations used for its within-fire evaluation. Transfer loss and differences between descriptors are therefore paired within fold before averaging. For whole-fire transfer AUC, uncertainty is estimated by resampling spatial blocks with replacement within the target folds using 2,000 bootstrap replicates. Because the regularisation selected within a source fire need not transfer best, we also select the probe regularisation by leaving out one source fire at a time and report this alongside the fixed probe.

\paragraph{Adaptation with target labels.}
Poor transfer can arise because the representation lacks relevant information for the target fire or because the rule learned elsewhere does not apply there. To separate these cases, we train on the three source fires plus $k \in \{0,2,4,8,16,32\}$ labelled blocks from the target fire, and compare this with a probe trained on the same $k$ target blocks alone. Target blocks are sampled from the target's training folds, preserving their spatial separation from the held-out test fold. Recovery with only limited target supervision indicates that the information is present in the representation but is not used correctly by the transferred rule.

\paragraph{Comparing learned rules.}
A fitted linear probe defines a direction in feature space through its weight vector. For this diagnostic analysis only, features are standardised jointly across the four fires so that the fitted vectors share a common coordinate system. For each target fire, $w_t$ is fitted on four of its folds and $s$ on the other three fires. We decompose $r = w_t - (w_t \cdot s)\,s,$ where the component along $s$ captures the direction shared with the other fires and $r$ the target-specific component. The held-out target fold is scored with $s$, $r$, and $w_t$ to quantify how much predictive information is carried by the shared and fire-specific directions.

\paragraph{Temporal generalization.}

The four benchmark fires all occurred in 2023 and therefore cannot separate geographic from temporal generalization. We use two additional fires in Attica, Greece: Parnitha, which burned in August 2023, and Varnavas, twelve kilometres away, which burned in August 2024. Parnitha is represented by 2022 descriptors and Varnavas by 2023 descriptors, so training on Parnitha and applying the probe to Varnavas uses only information available before the target fire and constitutes a forecasting setting.

Varnavas is evaluated on the same spatial folds and with the same paired loss as above. We compare transfer from Parnitha with transfer from each of the four fires in other regions, the four remote fires pooled, and Parnitha combined with the pooled remote fires. We additionally evaluate the reverse transfer from Varnavas to Parnitha and repeat the target-label adaptation experiment on Varnavas with Parnitha as the source.

\section{Results}
\label{sec:results}

\subsection{Frozen embeddings outperform conventional descriptors within fires}

Table~\ref{tab:within} gives the within-fire AUC for every fire and representation, and Figure~\ref{fig:folds} in Appendix~\ref{app:extra} shows the spread across allocations for the Greek fires. TESSERA leads the Sentinel-2 summaries by $+0.061$ to $+0.132$ and AlphaEarth by $+0.053$ to $+0.105$, and the paired differences are positive in nearly every allocation. Neither embedding is consistently ahead of the other. TESSERA is ahead on Rhodes by 0.08 in every allocation, the two are level on Evros, and AlphaEarth is ahead on C\'aceres and Tenerife by 0.03 and 0.01. Concatenating them gives the highest within-fire AUC on every fire, by $+0.006$ to $+0.047$ over the better single embedding, so locally they carry partly different information. The spread across allocations is not small: on Rhodes, single allocations of TESSERA range from 0.77 to 0.87, and allocation 0, the one the standard splitter produces, sits near the top of that range.

\begin{table}[t]
\caption{Within-fire ROC AUC, logistic regression, mean $\pm$ SD over 21 block-to-fold allocations, 79,056 pixels. TESSERA is the dequantised embedding. The lower rows are paired differences computed within fold, with the number of allocations in which the difference is positive. Bold: best representation per fire. Arrows give the direction of improvement.}
\label{tab:within}
\begin{center}
\small
\setlength{\tabcolsep}{5pt}
\begin{tabular}{lcccc}
\toprule
Representation & Rhodes $\uparrow$ & Evros $\uparrow$ & C\'aceres $\uparrow$ & Tenerife $\uparrow$ \\
\midrule
WorldCover & $0.629 \pm 0.014$ & $0.639 \pm 0.011$ & $0.597 \pm 0.028$ & $0.814 \pm 0.038$ \\
Sentinel-2 (15) & $0.693 \pm 0.028$ & $0.745 \pm 0.009$ & $0.763 \pm 0.044$ & $0.890 \pm 0.030$ \\
Sentinel-2, no counts & $0.681 \pm 0.019$ & $0.742 \pm 0.011$ & $0.750 \pm 0.047$ & $0.887 \pm 0.042$ \\
TESSERA & $0.825 \pm 0.026$ & $0.853 \pm 0.007$ & $0.824 \pm 0.061$ & $0.958 \pm 0.014$ \\
AlphaEarth & $0.746 \pm 0.033$ & $0.850 \pm 0.008$ & $0.851 \pm 0.035$ & $0.969 \pm 0.009$ \\
AlphaEarth + TESSERA & $\mathbf{0.832} \pm 0.027$ & $\mathbf{0.900} \pm 0.006$ & $\mathbf{0.861} \pm 0.032$ & $\mathbf{0.975} \pm 0.009$ \\
\midrule
TESSERA $-$ Sentinel-2 (15) & $+0.132$ (21/21) & $+0.108$ (21/21) & $+0.061$ (20/21) & $+0.068$ (21/21) \\
AlphaEarth $-$ Sentinel-2 (15) & $+0.053$ (19/21) & $+0.105$ (21/21) & $+0.088$ (21/21) & $+0.079$ (21/21) \\
AlphaEarth $-$ TESSERA & $-0.079$ (0/21) & $-0.002$ (7/21) & $+0.026$ (14/21) & $+0.012$ (19/21) \\
\bottomrule
\end{tabular}
\end{center}
\end{table}

\subsection{The within-fire advantage is robust}

\textbf{Spatial buffers and block bootstrap.} Removing training pixels within 1\,km of every test pixel lowers every representation by a similar amount, so the gap is unchanged, and at that buffer the block bootstrap puts the TESSERA minus Sentinel-2 difference at $+0.111$ (95\% interval $+0.014$ to $+0.182$) on Rhodes and $+0.087$ ($+0.060$ to $+0.111$) on Evros (Appendix~\ref{app:extra}).

\textbf{Classifier family.} Under boosting (Table~\ref{tab:clf}, Appendix~\ref{app:extra}), TESSERA still leads the indices by $+0.042$ to $+0.119$ in every fire, in 24 or 25 of 25 folds, and AlphaEarth by $+0.038$ to $+0.103$, so the linear probe is not what holds the index summaries back. An independent re-implementation on the Greek fires by one of the authors, with XGBoost and pooled AUC, gave the same gap and the same held-out result.

\textbf{Dequantisation.} Using the raw integers instead of the dequantised embedding changes the within-fire AUC by at most 0.006 and the held-out AUC by at most 0.03.

\subsection{The embedding advantage disappears across regions}

Table~\ref{tab:firelevel} and Figure~\ref{fig:firelevel} (Appendix~\ref{app:extra}) give, for each fire, the AUC obtained within the fire and the AUC obtained on the same spatial folds when the probe was trained on the other three. TESSERA loses 0.13 to 0.23 AUC, 0.18 on average. AlphaEarth loses 0.11 to 0.18, 0.15 on average. The Sentinel-2 index summaries without observation counts lose 0.02 to 0.09, 0.06 on average. Held out, the three end up in the same place. The mean AUC is 0.69 for TESSERA, 0.71 for AlphaEarth, and 0.71 for the indices. The paired difference between TESSERA and the indices is $-0.037$, $-0.044$ and $-0.038$ on Rhodes, Evros and Tenerife and $+0.035$ on C\'aceres, and for AlphaEarth $-0.027$, $-0.037$, $-0.038$ and $+0.084$. The block-bootstrap intervals exclude zero only on Evros and on C\'aceres. Concatenating the two embeddings, the best representation within every fire, is held out no better than either alone except on Tenerife. The correct summary is not that the index summaries are ahead in another region, but that the within-fire margin of 0.05 to 0.13 is lost there. The sign of the held-out difference depends on the fire, and on two fires out of four, its size is within the bootstrap uncertainty.

\begin{table}[t]
\caption{Fire-level summary on matched spatial folds. Within: probe trained inside the fire. Held out: probe trained on the other three fires, scored on the same folds of the same 21 allocations. Loss: within minus held out, paired within fold. S2 = Sentinel-2 index summaries without observation counts. Logistic regression is used here. TESSERA is dequantised. Bold: best of the three representations per fire and column. Arrows give the direction of improvement.}
\label{tab:firelevel}
\begin{center}
\small
\setlength{\tabcolsep}{4pt}
\begin{tabular}{lccc|ccc|ccc}
\toprule
& \multicolumn{3}{c|}{TESSERA} & \multicolumn{3}{c|}{AlphaEarth} & \multicolumn{3}{c}{S2 indices} \\
Held-out fire & within $\uparrow$ & held out $\uparrow$ & loss $\downarrow$ & within $\uparrow$ & held out $\uparrow$ & loss $\downarrow$ & within $\uparrow$ & held out $\uparrow$ & loss $\downarrow$ \\
\midrule
Rhodes & \textbf{0.825} & 0.595 & 0.230 & 0.746 & 0.606 & 0.140 & 0.681 & \textbf{0.632} & \textbf{0.049} \\
Evros & \textbf{0.853} & 0.674 & 0.179 & 0.850 & 0.681 & 0.170 & 0.742 & \textbf{0.718} & \textbf{0.024} \\
C\'aceres & 0.824 & 0.697 & 0.128 & \textbf{0.851} & \textbf{0.746} & 0.105 & 0.750 & 0.662 & \textbf{0.089} \\
Tenerife & 0.958 & 0.793 & 0.165 & \textbf{0.969} & 0.793 & 0.177 & 0.887 & \textbf{0.831} & \textbf{0.056} \\
\midrule
mean & \textbf{0.865} & 0.690 & 0.176 & 0.854 & 0.707 & 0.148 & 0.765 & \textbf{0.711} & \textbf{0.055} \\
\bottomrule
\end{tabular}
\end{center}
\end{table}

The pairwise designs, gradient boosting, and source-fire tuning give the same result with more noise (Appendix~\ref{app:transfer}). Boosting keeps held-out TESSERA far below its boosted within-fire AUC and lowers held-out AlphaEarth on every fire, and choosing the regularisation by leaving out a source fire changes the held-out AUC by at most 0.02.

\subsection{Limited local supervision recovers the embedding advantage}

\begin{figure}[t]
\begin{center}
\includegraphics[width=0.72\linewidth]{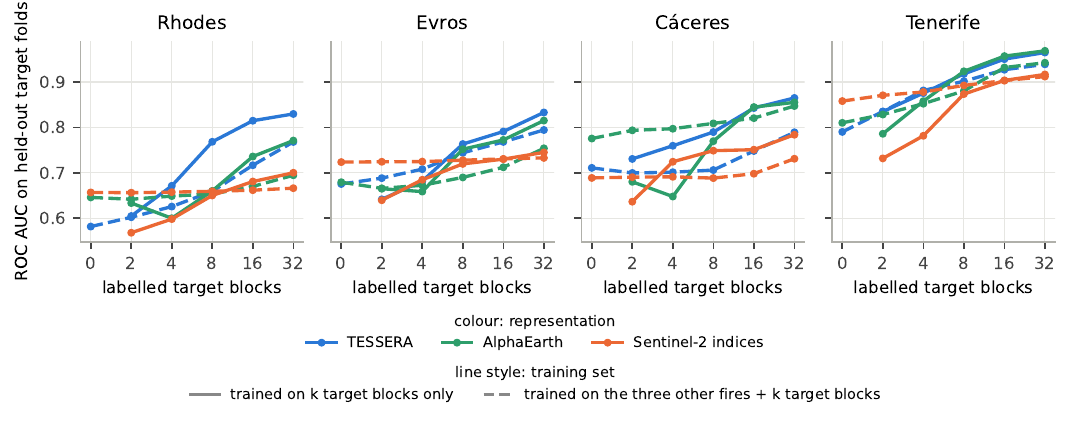}
\end{center}
\caption{Adaptation curves. AUC on the held-out target fold against the number of labelled target blocks in training, for probes trained on those blocks alone (solid) and on the three other fires plus those blocks (dashed). Mean over five folds and two allocations. The point at zero is the leave-one-fire-out probe.}
\label{fig:adapt}
\end{figure}

Figure~\ref{fig:adapt} shows how much can be recovered from labels from the target fire. With eight labelled blocks of the target, between 11 and 26\% of its training blocks, a probe trained on those blocks alone already puts TESSERA ahead of the indices trained the same way on every fire, by $+0.12$ on Rhodes and $+0.04$ on each of the others, and with sixteen blocks the margins are $+0.13$, $+0.06$, $+0.09$ and $+0.05$, most of the full within-fire advantage. AlphaEarth recovers similarly from sixteen blocks. Adding the three other fires, 59,000 labelled pixels, to the same target blocks makes the TESSERA probe worse than the target blocks alone from $k = 8$ on every fire, and from $k = 4$ on Rhodes and C\'aceres, by up to 0.10 AUC at $k = 16$. AlphaEarth behaves the same way from $k = 16$, and for the index summaries the other fires help at $k \leq 2$ and are neutral or mildly harmful by $k = 16$. So the embeddings contain what is needed to separate burned from control pixels in every fire. What does not carry from fire to fire is the linear rule, and a rule fitted on other fires actively interferes with one fitted locally.

\subsection{The learned linear rule is largely region-specific}

Table~\ref{tab:direction} in Appendix~\ref{app:extra} reports the split of each within-fire direction into the part shared with the other fires and the part specific to that fire. For the embeddings, the weight vector a probe learns inside a fire is nearly orthogonal to the one learned on the other three fires. The cosine is 0.12 to 0.22 for TESSERA and 0.03 to 0.41 for AlphaEarth, and 91 to 99\% of the within-fire coefficient norm lies in the fire-specific remainder. That remainder alone reproduces the full within-fire AUC, while the shared direction alone gives the held-out numbers of Table~\ref{tab:firelevel}. For the index summaries, the shared direction carries most of the within-fire AUC and the specific remainder adds little. The embeddings' local advantage is therefore not a sharper version of a rule shared by all fires. It is a different rule in each fire. The part of the embedding signal that does transfer is also low-dimensional and needs strong regularisation. Held out, a TESSERA probe on the top two to sixteen principal components of the training fires, or one regularised at $C = 10^{-4}$, gives a higher held-out AUC than the full-rank default probe on every fire (Figure~\ref{fig:rank}, Appendix~\ref{app:extra}), and choosing $C$ by leaving out a source fire, which selects $10^{-4}$ every time, raises the TESSERA held-out mean from 0.69 to 0.73, level with the indices, while within-fire AUC keeps rising to the full 128 dimensions.

\subsection{The embedding advantage is retained one year later}
\label{sec:attica}

We first apply the same within-fire evaluation to the two fires used for the temporal analysis. Across 21 spatial fold allocations, the AUC on Parnitha is $0.908 \pm 0.056$ for TESSERA, $0.945 \pm 0.050$ for AlphaEarth, and $0.749 \pm 0.090$ for the Sentinel-2 index descriptors. On Varnavas, the corresponding values are $0.809 \pm 0.092$, $0.831 \pm 0.078$, and $0.795 \pm 0.104$.

Transfer from Parnitha to Varnavas is strongest for TESSERA (Table~\ref{tab:attica}). It reaches an AUC of 0.772, only 0.040 below the within-Varnavas result, whereas training on a single fire from another region gives 0.628--0.721. AlphaEarth shows a weaker local advantage, with 0.740 from Parnitha and 0.758 from Tenerife. The Sentinel-2 index descriptors behave differently, reaching only 0.649 from Parnitha, compared with 0.790 from Evros and 0.757 from Rhodes. Pooling the four remote fires gives 0.765 for TESSERA and 0.717 for AlphaEarth, while combining Parnitha with the pooled fires gives the best transfer result for both embeddings, at 0.783 and 0.761. The adaptation analysis shows the same tendency for TESSERA: adding Parnitha improves performance with few Varnavas labels, whereas adding remote fires becomes harmful from eight target blocks onward (Figure~\ref{fig:adapt}).

The reverse transfer is weaker, with losses of 0.29 AUC for TESSERA and 0.16 for the Sentinel-2 index descriptors. Spatial and temporal proximity alone is therefore not sufficient, and the Attica pair should be treated as evidence of temporal transfer within one region rather than as a general property of the embeddings.

\begin{table}[t]
\caption{Transfer to the Varnavas 2024 fire from different training sources, evaluated on Varnavas's spatial folds (allocation 0, mean of five folds). Loss is the AUC difference relative to a probe trained within Varnavas. Bold indicates the best training source for each descriptor.}
\label{tab:attica}
\begin{center}
\small
\setlength{\tabcolsep}{5pt}
\begin{tabular}{lcc|cc|cc}
\toprule
& \multicolumn{2}{c|}{TESSERA} & \multicolumn{2}{c|}{AlphaEarth} & \multicolumn{2}{c}{S2 indices} \\
Training source & AUC $\uparrow$ & loss $\downarrow$ & AUC $\uparrow$ & loss $\downarrow$ & AUC $\uparrow$ & loss $\downarrow$ \\
\midrule
Rhodes, 2023 & 0.675 & 0.137 & 0.616 & 0.227 & 0.757 & 0.057 \\
Evros, 2023 & 0.628 & 0.183 & 0.697 & 0.146 & \textbf{0.790} & \textbf{0.025} \\
C\'aceres, 2023 & 0.629 & 0.183 & 0.672 & 0.171 & 0.638 & 0.176 \\
Tenerife, 2023 & 0.721 & 0.091 & 0.758 & 0.085 & 0.755 & 0.059 \\
\midrule
Parnitha, 12\,km away, 2023 & 0.772 & 0.040 & 0.740 & 0.103 & 0.649 & 0.165 \\
The four remote fires pooled & 0.765 & 0.046 & 0.717 & 0.126 & 0.772 & 0.043 \\
Parnitha and the four pooled & \textbf{0.783} & \textbf{0.029} & \textbf{0.761} & \textbf{0.082} & 0.771 & 0.043 \\
\midrule
within Varnavas & 0.812 &  & 0.843 &  & 0.814 &  \\
\bottomrule
\end{tabular}
\end{center}
\end{table}

\section{Discussion}
\label{sec:discussion}

This study evaluated whether frozen Earth observation embeddings can distinguish areas that later burn from those that do not, and how this ability generalizes across fires, regions, and time. Within fires, both embedding products consistently outperform the conventional descriptors: TESSERA exceeds the Sentinel-2 index summaries by 0.061--0.132 AUC and AlphaEarth by 0.053--0.105. Concatenating both embeddings gives the highest within-fire AUC for every fire, with a further gain of 0.006--0.047 over the better single embedding. These results remain stable across spatial allocations, buffering, classifier family, dequantisation, and nodata handling. Across regions, however, this advantage largely disappears. Strong within-fire performance therefore does not imply that a classifier fitted in one region transfers unchanged to another.

Much of the embedding advantage can nevertheless be recovered with limited local supervision. With only eight labelled blocks from the target fire, TESSERA already exceeds the index summaries trained on the same local data by 0.04--0.12 AUC, and with sixteen blocks much of the full within-fire advantage is recovered. Adding approximately 59,000 labelled pixels from the other fires can instead reduce performance relative to using the target blocks alone. This indicates that useful information remains accessible in the embeddings, but that the rule learned in other fires does not transfer well. Consistently, the fitted weight vectors are nearly orthogonal across fires, suggesting that the linear relationship between the embeddings and wildfire occurrence is largely fire-specific. The conventional descriptors show the opposite trade-off: they provide weaker within-fire discrimination, but the relationships learned from them transfer more consistently across regions.

This distinction is important for how frozen Earth observation embeddings are evaluated and used. A benchmark that evaluates a frozen representation only with a locally fitted head measures whether the representation contains information useful in that region, but not whether the fitted rule will generalize elsewhere. Our results suggest that these embeddings may be more useful as reusable representations for lightweight local models than as the basis for a single transferable classifier. This is also an attractive use of foundation models: the expensive representation has already been learned and released, while downstream applications can adapt a small model to local labels instead of training a new representation model for every task and region. This reuse does not by itself establish a lower environmental cost, but it reduces the need to repeatedly train large representation models from scratch.

The Attica experiment gives an initial indication that such local models may also remain useful over time. For TESSERA, a probe trained on Parnitha in 2023 reaches 0.772 AUC on Varnavas in 2024, only 0.040 below a probe trained within Varnavas itself. Individual fires from other regions reach only 0.628--0.721, while pooling the four remote fires reaches 0.765. AlphaEarth shows a weaker local effect, although combining Parnitha with the pooled remote fires gives the best AlphaEarth transfer result at 0.761. The Attica pair does not establish general temporal stability of the embedding model, but is a first indication that useful cross-year transfer within a region is possible.

This temporal setting also points to a practical direction for future work. In regions with recurrent wildfire risk and previously mapped fires, historical events could provide local supervision while annual foundation-model embeddings provide updated pre-fire descriptors for subsequent years. Region-specific models could then be used to rank areas by susceptibility before a fire occurs and potentially support the prioritisation of monitoring or preventive resources. Because the present study deliberately uses simple probes to test the information contained in the representations, this setting can be extended to richer downstream models without retraining the foundation model itself.

\section{Limitations}

The study is based on six fires, and the cross-regional fire-level comparisons therefore have a small effective sample size. The temporal analysis is particularly limited because it rests on a single pair of fires and is asymmetric between the two transfer directions. It cannot establish how transfer changes with distance, landscape similarity, or the number of years between source and target fires. A larger evaluation with repeated fires from the same regions over several years is therefore needed before the forecasting result can be generalized.

The fires are restricted to Mediterranean and Macaronesian environments and two consecutive years, and the conventional baseline is intentionally simple. Stronger baselines using additional Sentinel-2 bands, Sentinel-1 radar, terrain, weather, or other physical variables may reduce the observed embedding advantage. The control pixels also differ in composition between fires, so matched or land-cover-stratified controls would help determine how much of the cross-regional loss is compositional. Finally, the fire-specific part of the embedding signal remains unexplained; relating it to land cover, vegetation, terrain, climate, and fuel characteristics is an important next step. Practical sensitivities in grouped cross-validation and nodata handling are documented in Appendix~\ref{app:pitfalls}. Before these models could support operational fire management, they would need to be evaluated across substantially more regions and fire seasons.

\section{Conclusion}
Using six wildfire events, we evaluated frozen Earth observation embeddings as pre-fire descriptors for distinguishing areas that later burned from those that did not. The embeddings consistently outperform conventional descriptors in within-region evaluation, but this advantage largely disappears under cross-regional generalization. Limited local supervision recovers much of the embedding advantage, indicating that useful information is present in the representations while the fitted linear rule is strongly region-specific. The temporal analysis further suggests that region-specific models can remain useful across years, motivating broader evaluation of temporal generalization for wildfire forecasting.

%Inside each of four 2023 wildfires, two independently built frozen embeddings separated burned from unburned pixels clearly better than conventional index summaries did, and that advantage was unchanged under all our robustness checks. On a fire in another region, that advantage was lost, however. A few labelled blocks from the new region, however, restored it. Labels from other regions lowered the AUC further, and the weight vector learned inside one region was nearly orthogonal to the vector learned in the others. In the one experiment that is a forecast, a probe fitted on a 2023 fire and applied to a fire twelve kilometres away in 2024 lost 0.04 AUC, so the probe transfers across one year of elapsed time but not across a change of landscape. Evaluation inside one region cannot distinguish an advantage that carries over to a new region from one that does not, and the three steps used here can, with only a linear probe and the labels one already has. Studies that use frozen embeddings as descriptors should report held-out-region performance as a primary result, save their fold assignments with the code, and check their inputs for nodata vectors.

\subsection*{Reproducibility statement}

Every number in the paper is read from a committed output table produced by a script in the accompanying repository. (Not available in this preprint version.) The repository contains the sampling, extraction and analysis code, the 120,000-row pixel table for the six fires with labels, coordinates, block identifiers and all four descriptor sets, the per-pixel TESSERA scale values, the committed fold table for all 21 allocations and four fires, out-of-fold probabilities for every probe, and the environment lock file. CEMS products, WorldCover tiles, and TESSERA tiles are downloaded by a preparation script from their public hosts with recorded checksums (Appendix~\ref{app:data}). TESSERA v1.0 tiles are served from the \texttt{tessera-embeddings} S3 bucket, since the host we first downloaded them from has been retired. AlphaEarth values are sampled from Earth Engine by a script in the repository, and the sampled values are committed so that the analysis does not require an Earth Engine account.

\subsection*{AI use statement}

Generative AI tools were used to assist with some language editing as well as improving some code mainly for clarity. All AI-assisted outputs were critically reviewed, tested, and validated by the authors. The authors take full responsibility for the final content of the manuscript, including its code, results, figures, claims, and references.

\subsection*{Ethics statement}

The study uses only public satellite products and public emergency-mapping polygons at 10\,m resolution and involves no personal data. The probes are research models. They have not been validated for operational fire-risk assessment and should not be used for it.

%\subsection*{Acknowledgements}
%Will be filled in after the double-blind review process is finished.
%This research was supported by the Swedish Research Council for Sustainable Development (FORMAS), grant no.\ 2022-151862. The research of A.S.\ is also supported by VR TWAR, grant no.\ 162385. Computational resources were provided by LUNARC, the centre for scientific and technical computing at Lund University.

\bibliography{iclr2027_conference}
\bibliographystyle{iclr2027_conference}

\appendix

\section{Data and processing details}
\label{app:data}

\begin{table}[h]
\caption{The six wildfire events. The first four are the benchmark. The last two are the Attica pair of Section~\ref{sec:attica}, with descriptors from 2022 and 2023 respectively. Blocks are 5\,km spatial cross-validation blocks. No-data pixels are control pixels with no TESSERA embedding (all-zero vector), nearly all permanent water. They are excluded from all analyses. Sentinel-2 observations are the median number of cloud-free pre-fire observations per pixel.}
\label{tab:events}
\begin{center}
\small
\setlength{\tabcolsep}{4pt}
\begin{tabular}{llllrrrr}
\toprule
Activation & Fire & Date & CEMS product & Burned (ha) & Blocks & Nodata & S2 obs. \\
\midrule
EMSR675 & Rhodes, Greece & Jul 2023 & DEL\_MONIT08 & 17,770 & 48 & 440 & 122 \\
EMSR686 & Evros, Greece & Aug 2023 & DEL\_MONIT09 & 93,939 & 90 & 338 & 122 \\
EMSR667 & C\'aceres, Spain & May 2023 & DEL\_PRODUCT v2 & 9,595 & 47 & 0 & 228 \\
EMSR685 & Tenerife, Spain & Aug 2023 & DEL\_MONIT02 & 10,960 & 39 & 162 & 58 \\
EMSR690 & Parnitha, Greece & Aug 2023 & DEL\_MONIT03 & 6,190 & 29 & 0 & 148 \\
EMSR746 & Varnavas, Greece & Aug 2024 & DEL\_MONIT02 & 10,405 & 20 & 0 & 128 \\
\bottomrule
\end{tabular}
\end{center}
\end{table}

\paragraph{CEMS products.} EMSR675 \texttt{EMSR675\_AOI01\_DEL\_MONIT08\_v1}, EMSR686 \texttt{EMSR686\_AOI01\_DEL\_MONIT09\_v1}, EMSR667 \texttt{EMSR667\_AOI01\_DEL\_PRODUCT\_v2}, EMSR685 \texttt{EMSR685\_AOI01\_DEL\_MONIT02\_v1}, EMSR690 \texttt{EMSR690\_AOI01\_DEL\_MONIT03\_v1} and EMSR746 \texttt{EMSR746\_AOI01\_DEL\_MONIT02\_v1}. The \texttt{observedEventA} layer gives burned polygons, and \texttt{areaOfInterestA} the activation area from which controls are drawn. Two of the 112 Evros polygons were topologically invalid on read and were repaired without change of area.

\paragraph{Sampling.} Interior buffer 30\,m, exclusion buffer 500\,m, control ratio 3, caps 5,000 and 15,000, seed 42. Blocks are 5\,km cells of the EPSG:3035 grid. Pixels are deduplicated by tile ownership at TESSERA tile boundaries.

\paragraph{Sentinel-2.} Level-2A from the Microsoft Planetary Computer STAC, all scenes of the pre-fire year over each activation bounding box (438 Rhodes, 334 Evros, 635 C\'aceres, 99 Tenerife, 337 Parnitha, 234 Varnavas). Reflectance is $(\mathrm{DN} - 1000)/10000$. B04 and B08 nearest at 10\,m, B11 and B12 bilinear from 20\,m. Pixels with no valid observation (four on Tenerife) are dropped.

\paragraph{TESSERA.} v1.0, variant \texttt{vultr}, year 2022 for the five 2023 fires and year 2023 for the Varnavas fire, 58 tiles for the benchmark and 19 more for the Attica pair. Embedding $= \mathrm{int8} \times \mathrm{scale}$, where scale is a float32 per pixel in the range 0.04 to 0.12 over our pixels. Fold seeds and classifier settings are in the repository.

\paragraph{AlphaEarth.} Earth Engine image collection \path{GOOGLE/SATELLITE_EMBEDDING/V1/ANNUAL}, bands A00 to A63, sampled at each pixel centre at 10\,m without resampling, for the pre-fire year of each activation. Values lie in $[-0.48, 0.50]$ and every sampled vector has unit norm. The dataset is released under CC BY 4.0 and is produced by Google and Google DeepMind.

\section{Additional within-fire results}
\label{app:extra}

\begin{figure}[t]
\begin{center}
\includegraphics[width=0.6\linewidth]{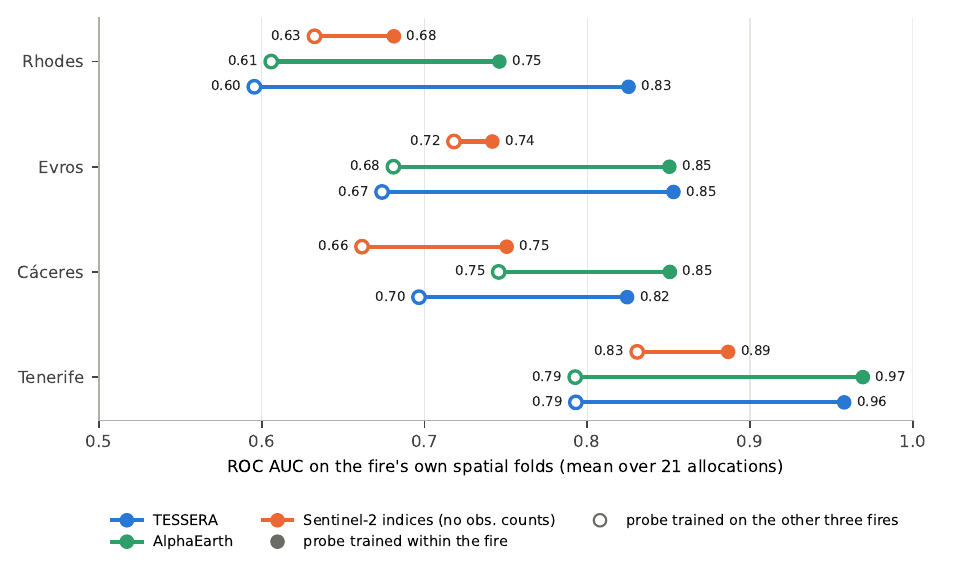}
\end{center}
\caption{AUC on each fire's own spatial folds when the probe is trained within the fire (filled) and when it is trained on the other three fires (open), for TESSERA, AlphaEarth, and the Sentinel-2 index summaries.}
\label{fig:firelevel}
\end{figure}

\begin{figure}[h]
\begin{center}
\includegraphics[width=0.9\linewidth]{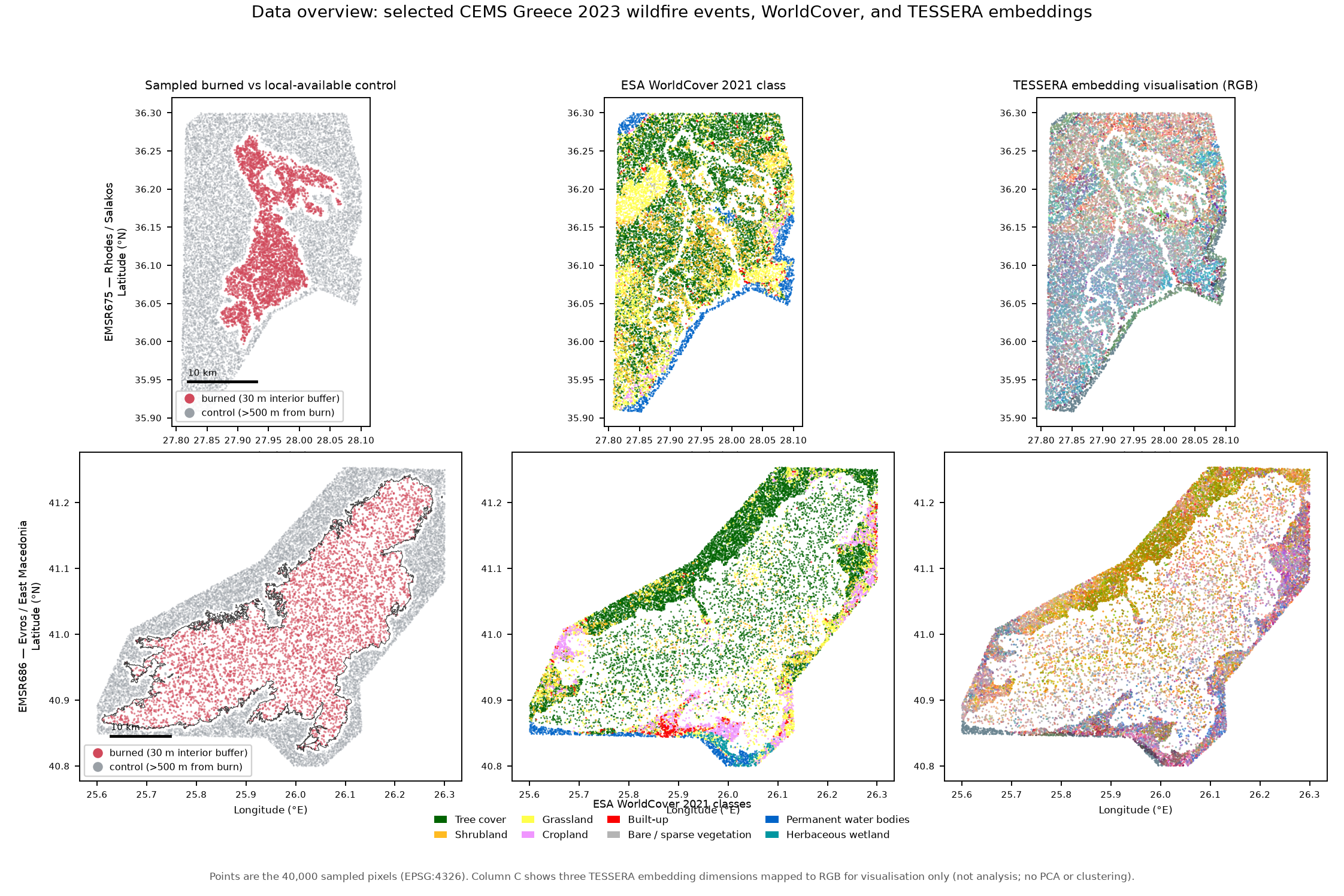}
\end{center}
\caption{Sampled burned and control pixels, WorldCover class, and a three-dimensional TESSERA visualisation for Rhodes (top) and Evros (bottom).}
\label{fig:data-overview}
\end{figure}

\begin{figure}[h]
\begin{center}
\includegraphics[width=\linewidth]{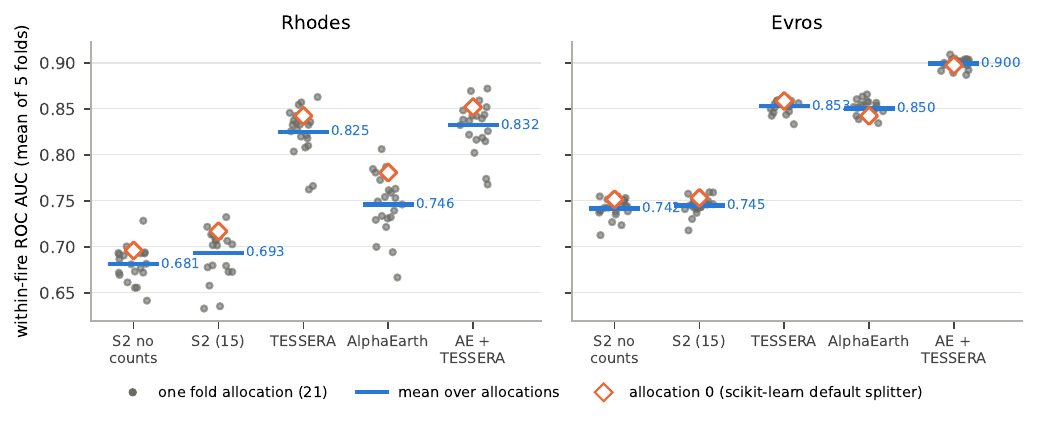}
\end{center}
\caption{Within-fire AUC under each of 21 block-to-fold allocations for the two Greek fires. Grey points are single allocations, bars are their mean, and diamonds are allocation 0, the one produced by scikit-learn's default splitter.}
\label{fig:folds}
\end{figure}

\begin{table}[h]
\caption{Within-fire ROC AUC with a tuned logistic regression (LR) and with gradient boosting (GBM) on the same persisted folds (allocations 0 to 4, mean over folds and allocations, TESSERA raw integers, which differ from the dequantised embedding by at most 0.006 within fire). The last two columns are the paired TESSERA minus Sentinel-2 (15) difference under each classifier. Bold: best representation per fire under each classifier.}
\label{tab:clf}
\begin{center}
\small
\setlength{\tabcolsep}{4pt}
\begin{tabular}{lcccccccccc}
\toprule
& \multicolumn{2}{c}{Sentinel-2 (15)} & \multicolumn{2}{c}{S2 no counts} & \multicolumn{2}{c}{TESSERA} & \multicolumn{2}{c}{AlphaEarth} & \multicolumn{2}{c}{$\Delta$ (T $-$ S2)} \\
Fire & LR $\uparrow$ & GBM $\uparrow$ & LR $\uparrow$ & GBM $\uparrow$ & LR $\uparrow$ & GBM $\uparrow$ & LR $\uparrow$ & GBM $\uparrow$ & LR & GBM \\
\midrule
Rhodes & 0.700 & 0.705 & 0.691 & 0.706 & \textbf{0.821} & \textbf{0.824} & 0.734 & 0.743 & $+0.120$ & $+0.119$ \\
Evros & 0.748 & 0.784 & 0.748 & 0.759 & \textbf{0.848} & 0.858 & 0.846 & \textbf{0.877} & $+0.100$ & $+0.074$ \\
C\'aceres & 0.774 & 0.791 & 0.758 & 0.767 & 0.823 & 0.856 & \textbf{0.866} & \textbf{0.894} & $+0.048$ & $+0.064$ \\
Tenerife & 0.891 & 0.926 & 0.885 & 0.923 & 0.955 & 0.968 & \textbf{0.968} & \textbf{0.976} & $+0.065$ & $+0.042$ \\
\bottomrule
\end{tabular}
\end{center}
\end{table}

\begin{table}[h]
\caption{Within-fire probe direction decomposed along the direction learned on the other three fires (allocation 0, mean of five folds, global standardisation). cos: cosine between the two directions. AUC on the test fold using the shared direction alone, the fire-specific remainder alone, and the full within-fire direction. Bold: larger of shared and specific.}
\label{tab:direction}
\begin{center}
\small
\setlength{\tabcolsep}{3.5pt}
\begin{tabular}{lcccc|cccc|cccc}
\toprule
& \multicolumn{4}{c|}{TESSERA} & \multicolumn{4}{c|}{AlphaEarth} & \multicolumn{4}{c}{S2 indices} \\
Fire & cos & shared & specific & full & cos & shared & specific & full & cos & shared & specific & full \\
\midrule
Rhodes & 0.12 & 0.594 & \textbf{0.837} & 0.843 & 0.20 & 0.605 & \textbf{0.770} & 0.781 & 0.54 & \textbf{0.634} & 0.625 & 0.697 \\
Evros & 0.20 & 0.674 & \textbf{0.847} & 0.859 & 0.41 & 0.678 & \textbf{0.747} & 0.843 & 0.24 & 0.719 & \textbf{0.746} & 0.751 \\
C\'aceres & 0.22 & 0.699 & \textbf{0.859} & 0.869 & 0.03 & 0.752 & \textbf{0.868} & 0.870 & $-0.25$ & 0.676 & \textbf{0.765} & 0.767 \\
Tenerife & 0.16 & 0.817 & \textbf{0.955} & 0.957 & 0.09 & 0.793 & \textbf{0.957} & 0.959 & 0.71 & \textbf{0.855} & 0.831 & 0.912 \\
\bottomrule
\end{tabular}
\end{center}
\end{table}

\paragraph{Classifier grids.} The tuned logistic regression searches $C \in \{0.01, 0.1, 1, 10\}$. LightGBM uses 300 trees, learning rate 0.05, leaves $\in \{15, 63\}$, minimum leaf size $\in \{20, 100\}$ and balanced weights. Boosting lifts the index summaries by $+0.005$ to $+0.036$, TESSERA by $+0.003$ to $+0.033$ and AlphaEarth by $+0.008$ to $+0.031$, and on the held-out set it reaches 0.63, 0.65, 0.69 and 0.86 for TESSERA against 0.60, 0.70, 0.65 and 0.85 for the indices.

\paragraph{Buffers.} Removing training pixels within 1\,km of every test pixel, about 14\% of the training set, lowers TESSERA on allocation 0 of the Greek fires from 0.846 to 0.820 (Rhodes) and from 0.855 to 0.839 (Evros), with Sentinel-2 at 0.709 and 0.753 under the same buffer. Held out, dequantisation helps by $+0.02$ on Evros and $+0.03$ on Tenerife, with bootstrap intervals excluding zero, and by less than 0.01 elsewhere.

\paragraph{Tenerife is the easiest fire, and that is informative.}

Tenerife has the highest AUC for every representation, including 0.81 for the land-cover class alone. Its burned area is almost entirely pine forest, and its controls are a mixture of shrub, bare lava, built-up land, and coast, so the burned--control contrast is largely a land-cover contrast. It is also the easiest target for probes trained on Evros (0.75 to 0.85 for every representation) and the worst source (0.49 to 0.71 on the other fires). The transfer numbers reflect how alike the burned--control contrasts of two fires are, not how much a representation knows about fire.

\begin{figure}[h]
\begin{center}
\includegraphics[width=\linewidth]{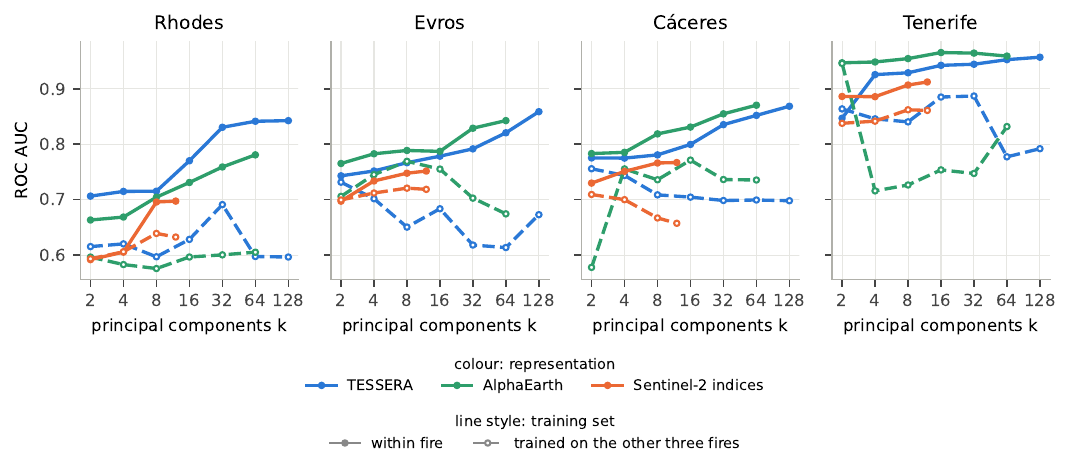}
\end{center}
\caption{AUC of the probe on the top $k$ principal components, within-fire (PCA fitted on the training folds) and held-out (PCA fitted on the three training fires), allocation 0. Within-fire AUC of the embeddings keeps rising to full rank. Held-out AUC peaks at a few components.}
\label{fig:rank}
\end{figure}

\section{Full transfer results}
\label{app:transfer}

\paragraph{Pairwise transfer, boosting and source-fire tuning.} The pairwise designs give the same result as the leave-one-fire-out designs, with more noise. Across the twelve directions, the indices are ahead of TESSERA in five and behind in seven, and ahead of AlphaEarth in eight. The largest gaps in either direction involve Tenerife as source or target (C\'aceres to Tenerife gives TESSERA 0.68 and the indices 0.53. Rhodes to Tenerife gives the indices 0.83 and AlphaEarth 0.59). Going from one training fire to three raises the mean held-out AUC of TESSERA by about 0.01 and of AlphaEarth by about 0.08 and does not restore the within-fire margin, though four fires cannot say what many would do. Choosing the probe's regularisation by leaving out a source fire changes the held-out AUC by at most 0.02 in either direction. Gradient boosting changes the order of the representations but not the conclusion. Held out, boosted TESSERA reaches 0.63, 0.65, 0.69 and 0.86 against 0.60, 0.70, 0.65 and 0.85 for the boosted indices, ahead on three fires by 0.01 to 0.04 but far below its boosted within-fire AUC of 0.82 to 0.97, while boosted AlphaEarth falls to 0.55, 0.66, 0.68 and 0.81, below its linear probe on every fire. One failure is worth explaining. The full Sentinel-2 set, including observation counts, trained with boosting on the other fires, scores 0.34 on C\'aceres, well below chance, because C\'aceres has twice the cloud-free coverage of the Greek fires and the trees split on the count columns. Observation counts are a property of the acquisition, not the surface, and should not be in a baseline that is meant to transfer.

\paragraph{The Attica pair.} Whole-fire AUC, which ranks all pixels of the target together instead of averaging over folds, gives the same picture as Table~\ref{tab:attica}. From Parnitha to Varnavas, TESSERA reaches 0.738, AlphaEarth 0.719, the index summaries 0.669, and the two embeddings concatenated 0.754. In the reverse direction, from Varnavas to Parnitha, TESSERA reaches 0.619, AlphaEarth 0.695, and the index summaries 0.593, against within-fire values of 0.91, 0.95, and 0.75 on Parnitha. The embeddings therefore stay ahead of the index summaries in both directions of this pair, which happens for no remote pair, but only the forward direction keeps most of the within-fire performance.

\begin{table}[h]
\caption{Pairwise and leave-one-fire-out transfer ROC AUC over the whole target fire, logistic regression, nodata excluded. Whole-fire AUCs differ from the fold-averaged values of Table~\ref{tab:firelevel} because they also rank pixels across folds. Bold: best representation per row. S2 (15) includes observation counts and S2 idx excludes them. T. deq. = TESSERA dequantised, AE = AlphaEarth, AE+T = AlphaEarth and TESSERA concatenated.}
\label{tab:pairwise}
\begin{center}
\footnotesize
\setlength{\tabcolsep}{2.2pt}
\begin{tabular}{llccccccc}
\toprule
Train & Test & S2 (15) $\uparrow$ & S2 idx $\uparrow$ & TESSERA $\uparrow$ & T. deq. $\uparrow$ & AE $\uparrow$ & AE+T $\uparrow$ \\
\midrule
Rhodes & Evros & 0.640 & \textbf{0.663} & 0.654 & 0.642 & 0.567 & 0.639 \\
Rhodes & C\'aceres & 0.526 & 0.604 & 0.620 & 0.617 & 0.684 & \textbf{0.732} \\
Rhodes & Tenerife & \textbf{0.847} & 0.828 & 0.777 & 0.786 & 0.589 & 0.678 \\
Evros & Rhodes & 0.562 & \textbf{0.632} & 0.568 & 0.563 & 0.553 & 0.495 \\
Evros & C\'aceres & \textbf{0.717} & 0.702 & 0.676 & 0.654 & 0.663 & 0.684 \\
Evros & Tenerife & 0.847 & 0.849 & 0.753 & 0.766 & 0.826 & \textbf{0.873} \\
C\'aceres & Rhodes & 0.489 & 0.572 & 0.582 & 0.596 & 0.606 & \textbf{0.641} \\
C\'aceres & Evros & 0.682 & 0.678 & 0.681 & \textbf{0.694} & 0.621 & 0.689 \\
C\'aceres & Tenerife & 0.511 & 0.527 & 0.684 & \textbf{0.732} & 0.582 & 0.648 \\
Tenerife & Rhodes & 0.524 & 0.626 & 0.650 & \textbf{0.653} & 0.494 & 0.546 \\
Tenerife & Evros & 0.689 & 0.706 & 0.686 & \textbf{0.712} & 0.666 & 0.688 \\
Tenerife & Cáceres & 0.683 & 0.678 & 0.689 & 0.696 & 0.678 & \textbf{0.700} \\
\midrule
Evros + C\'aceres + Tenerife & Rhodes & \textbf{0.660} & 0.632 & 0.595 & 0.596 & 0.605 & 0.606 \\
Rhodes + C\'aceres + Tenerife & Evros & \textbf{0.719} & \textbf{0.719} & 0.655 & 0.673 & 0.675 & 0.671 \\
Rhodes + Evros + Tenerife & C\'aceres & 0.629 & 0.658 & 0.700 & 0.699 & 0.735 & \textbf{0.741} \\
Rhodes + Evros + C\'aceres & Tenerife & 0.864 & 0.861 & 0.766 & 0.792 & 0.832 & \textbf{0.915} \\
\bottomrule
\end{tabular}
\end{center}
\end{table}

\begin{table}[h]
\caption{Leave-one-fire-out transfer ROC AUC with gradient boosting, tuned inside the training fires. Bold: best representation per row.}
\label{tab:lofo-gbm}
\begin{center}
\small
\begin{tabular}{lccccc}
\toprule
Held-out fire & S2 (15) $\uparrow$ & S2 idx $\uparrow$ & TESSERA $\uparrow$ & TESSERA deq. $\uparrow$ & AlphaEarth $\uparrow$ \\
\midrule
Rhodes & \textbf{0.645} & 0.601 & 0.630 & 0.632 & 0.550 \\
Evros & 0.638 & \textbf{0.697} & 0.651 & 0.668 & 0.658 \\
C\'aceres & 0.342 & 0.648 & \textbf{0.687} & 0.681 & 0.680 \\
Tenerife & 0.798 & 0.852 & 0.859 & \textbf{0.861} & 0.806 \\
\bottomrule
\end{tabular}
\end{center}
\end{table}

\section{Two pitfalls we met - beyond this study}
\label{app:pitfalls}

We also report two practical problems we encountered and which we believe could be valuable to know for the community beyond this study alone. First, scikit-learn's \texttt{StratifiedGroupKFold} only shuffles groups that have similar class ratios, as its documentation states. With a few dozen blocks per fire, almost every random seed then produces the same split into folds, and for Rhodes, 21 seeds gave only 3 distinct splits, ten of them identical, so a standard deviation computed over seeds is not a measure of split-to-split variability. We drew our own random block-to-fold allocations instead, saved all 21 of them in the repository, and used that saved table for every experiment. Second, TESSERA marks pixels without an embedding by an all-zero vector. In our sample, these are almost all permanent water, so they occur only among the controls, and a classifier can learn ``all zeros means control'' without learning anything about the land surface. The effect here was at most 0.004 AUC, but it grows with class imbalance, and a cross-validation that leaves such pixels in cannot detect it.

\section{Descriptive results for the Greek fires}
\label{app:descriptive}

Burned pixels in the Greek fires are over-represented in tree cover (0.683 of burned versus 0.517 of control pixels) and shrubland (0.162 versus 0.087) and under-represented in grassland and cropland. In the 128-dimensional embedding space, the burned and control centroids are separated (Euclidean distance 76.9 pooled, on raw integer values), and the separation persists within every WorldCover class with at least 100 burned and 100 control pixels, so the embedding contrast is not a restatement of land-cover composition. The full tables are in the repository.

\end{document}